# OIL AND GAS PIPELINE MONITORING DURING COVID-19 PANDEMIC VIA UNMANNED AERIAL VEHICLE

**Myssar Jabbar Hammood AL-BATTBOOTTI, Iuliana MARIN, Nicolae GOGA, Ramona POPA**

*University Politehnica of Bucharest, Faculty of Engineering in Foreign Languages (ROMANIA), marin.iulliana25@gmail.com*

## Abstract

The vast network of oil and gas transmission pipelines requires periodic monitoring for maintenance and hazard inspection to avoid equipment failure and potential accidents. The severe COVID-19 pandemic situation forced the companies to shrink the size of their teams. One risk which is faced on-site is represented by the uncontrolled release of flammable oil and gas. Among many inspection methods, the unmanned aerial vehicle system contains flexibility and stability. Unmanned aerial vehicles can transfer data in real-time, while they are doing their monitoring tasks. The current article focuses on unmanned aerial vehicles equipped with optical sensing and artificial intelligence, especially image recognition with deep learning techniques for pipeline surveillance.

Unmanned aerial vehicles can be used for regular patrolling duties to identify and capture images and videos of the area of interest. Places that are hard to reach will be accessed faster, cheaper and with less risk. The current paper is based on the idea of capturing video and images of drone-based inspections, which can discover several potential hazardous problems before they become dangerous. Damage can emerge as a weakening of the cladding on the external pipe insulation. There can also be the case when the thickness of piping through external corrosion can occur. The paper describes a survey completed by experts from the oil and gas industry done for finding the functional and non-functional requirements of the proposed system.

Keywords: Pipeline, image processing, thermography, unmanned aerial vehicle.

## 1   INTRODUCTION

Worldwide companies are greatly affected by the COVID-19 pandemic, thus the new situation leads to the reduction of either work or staff, especially in the oil and gas fields. As people start to work at home, the staff presence decreases, while accidents, theft, and equipment failure begin to show up in transmission lines. Any spill or leakage may lead to the creation of an environmental disaster. The risk of this industry demands close and periodic monitoring of its systems, such as pipelines and refineries.

The current paper proposes an aerial monitoring solution based on software to control and analyses captured data via unmanned aerial vehicle (UAV) over pipelines and sites. Firstly, weekly monitoring inspection and advanced technologies in sensing and imaging present a wide range of settings to carry out predictive maintenance of pipelines and merging UAV data with analytics. Secondly, this can help to predict the functionality status of pipes and identify defects, and to reduce the rate of unplanned shutdown failure as well as potential malfunctions. When conducting inspections manually, often several days or months are required for one inspection operation which means much more time to check the condition of a pipeline network. Thus, a shutdown means a significant loss of revenue for a company. Finally, by reducing the inspection time and avoiding the inspector's team, a dangerous situation can be bypassed by 70-80% which represents a massive potential saving in costs and hazards.

The article is quantitative research examining functional and non-functional requirements for drone surveillance of oil and gas pipelines. The main benefits are represented by risk reduction, namely inspection teams avoid potentially dangerous situations. Moreover, good records are obtained via the drone's meticulous record of the asset condition over time. The assets are smokestacks, drilling equipment, storage tanks and refineries. Another benefit is represented by less downtime. Manual inspections usually require cut production and can be a significant expense. By replacing this operation with drone inspections, it allows companies to acquire a full detailed view of potential damages. Time and costs can also be reduced. In addition to reducing expenses, drone inspections can reduce involved working hours.

The design of the current paper is as follows. Section 2 outlines the proposed system. Section 3 details the basic steps of inspection. Section 4 describes the results based on the survey answered by 77 professionals. The last one includes conclusions and future work.

## 2   LITERATURE REVIEW

Unmanned aerial vehicles (UAVs) are used in several monitoring fields due to their high efficiency and accuracy in analyzing data. A plethora of studies was done in different inspection fields using UAVs. Below is the review for some of these studies:

*Table 1. Studies regarding UAVs*

| Closely related work | Strength | Weakness | Our contribution |
|---|---|---|---|
| A UAV SAR Prototype for Marine and Arctic Application [1] | 1. Synthetic Aperture Radar (SAR) systems create high-resolution images. 2. The system works in various weather conditions, as well as during night or day to detect spills. | 1. High costs due to the technologies which are mounted inside an airplane or satellite. 2. The designed system is mainly used to detect crude offshore oil spills. | 1. Affordable sensors which will operate in any mission. 2. Several types of pipeline damages and environments will be analyzed based on images and videos generated by vision sensors. |
| Distributed Operation of Collaborating Unmanned Aerial Vehicles for Time-Sensitive Oil Spill Mapping [2] | 1. The system offered anomalies identification, stochastic occupancy grid mappings, and UAV waypoints. 2. A test team worked to operate an oil system via a shoreline platform. | 1. The system was tested on only 10 images. 2. The oil application used to detect offshore oil spills showed a weak mapping accuracy of 55%. | 1. Multiple images will be used based on the captured frames. 2. The accuracy of our proposed system will be in the range of 80-95%, based on different test conditions. |
| Object Recognition in Aerial Images Using Convolutional Neural Networks [3] | 1. CNN algorithm for object identification and classification based on aerial images had an accuracy of 97.8%. 2. Ability to recognize object classes in real time UAV video feed, with an accuracy of 84%. | 1.GPS does not detect objects in real time. 2. The dataset which was used in training contained only 267 images. | 1. Drones will detect problems. They will announce the staff regarding the problems that can appear. 2. Multiple images and videos will be processed, optimizing surveillance costs and save time. |
| Drone Aided Machine-Learning Tool for Post-Earthquake Bridge Damage Reconnaissance [4] | 1.Structural dataset contained a total of 5,911 images regarding damaged and undamaged structures. 2. The entire inspection was completed within 20 minutes. | 1. The system only recognized some parts of the cracking and failed to detect a long crack when observed from outside the bounding box. 2. The accuracy of the spalling detection was 63%. | 1. Multiple problems which can appear will be detected, increasing the performance and quality of monitored pipelines. 2. The assessment accuracy of our system will belong to the range 80-95%. |
| Oil Spill Detection Using Machine Learning and Infrared | 1. The system is able to detect oil spills during nighttime by using infrared cameras. | 1. The designed system is mainly used to detect crude offshore oil spills or inside a port | 1. Our system will offer a complete inspection and monitoring service for the oil and gas |

| Closely related work | Strength | Weakness | Our contribution |
|---|---|---|---|
| Images [5] | 2. The system was developed based on the Keras framework. Keras is a high-level Python programming environment that uses Tensorflow as a backend. | environment. 2. The trained model works only with the infrared camera, while the research is done using a drone with two cameras (RGB and thermal infrared). | industry. 2. Different sensors will be placed on the monitoring drones, such that an extended analysis of the environment can be performed. |
| Thermal Infrared and Visual Inspection of Photovoltaic Installations by UAV Photogrammetry—Application Case: Morocco [6] | 1. The inspection of photovoltaic installations was done successfully using UAVs. 2. Errors were obtained at centimetric and millimetric level. | 1. The procedure was done in a semi-automatic manner in order to extract visual defects. | 1. The proposed solution will be fully autonomous, without the need of human involvement. |
| IR Thermography from UAVs to Monitor Thermal Anomalies in the Envelopes of Traditional Wine Cellars: Field Test [7] | 1. Semi-buried wine cellars were monitored based on thermal sensors placed on drones that surveilled the outdoor environment. 2. Roof cellar observation lead to the detection of roof tiles replacement and water infiltration because of building wall cracks. 3. The drone was useful for qualitative analysis of wine preservation. | 1. Difficulties were detected when surface temperature measurements were performed under real operating conditions. 2. The view angle greatly determines the radiometric captured data and it can lead to disturbances caused by specular reflections. | 1. After testing the drone system in laboratory conditions, we will extend to testing it in real operating situations. 2. The best angles which determine problems will be determined based on multiple trials and according to the suggestions of the experts from the oil and gas domains. |
| The Pro and Cons of Drone Thermography Methodology for Solar Plants [8] | 1. Thermographic inspection is used for the operation and maintenance of solar plants in order to identify and report anomalies, like the deterioration of electrical performance. 2. Prioritize the mitigation of issues to claim warranties. | 1. Orthomosaic thermographic images are created by collecting images when the drone is found at a high altitude and it is needed to stitch the resulting images, such usually fails, because the camera was too close to the solar panels. 2. Abnormal temperature gradient is hard to record accurately and this leads to the impossibility to detect more subtle problems. | 1. We will use sensors which are able to calibrate. Smaller focal length, like 9 mm, will be used for lenses to increase image footprints. 2. Radiometrically calibrated cameras will enable the capture of absolute temperature for every pixel. 3. Drone speed will be adjusted for motion blurs to be avoided. 4. Image acquisition will be based on overlapping, reaching 90% for front and side image overlaps. |

The current research serves as a complete inspection and monitoring service for the oil and gas industry. Starting from mission planning, which is done by an operator before any mission, to the collection of various data from different sensors, the research aims to increase the level of efficiency of maintenance. The detections of several types of damages in pipeline and facility environments will be done based on images and videos created by vision sensors.

A brief overview of several remote sensing technologies that are used to detect oil spills in the ocean is given in the related work table [1]. These costly technologies are perfectly mounted in airplanes or satellites, but, when used in a port for operational purposes, there are other aspects to be considered. For UAVs, these factors are cost, availability, and applicability. Another study [2] was based on a distributed manner to map offshore oil spills by using drones. Based on the created occupancy grid framework, UAVs were used to explore a knowledge domain. Three components of the model were implemented, namely anomaly detection, stochastic occupancy grid mapping, and UAV waypoint planning. The model was tested through the analysis of ten oil spill images and succeeded in identifying 55% of the oil spills.

For this study [4] the authors focus on this research to determine the damages After the seismic strike, they studied inspections of structural damages to optimize the emergency management, they used drones with artificial intelligence, that performed the destruction classification task. The authors used such technologies to inspect bridges in New Zealand's by creating a machine-learning model for destruction detection based on image analysis. The training of the algorithm was done to recognize cracks in concrete. The research reached up to 84.7% for damage detection and 63% for cracking and spalling, respectively. While the authors propose a new approach in this study [5], A UAV and an infrared camera are used to detect oil spills inside a port area. An oil spill can be detected and cleaned more efficiently and at a lower cost by implementing this solution. Also, there is a study done in Morocco [6] by involving UAVs to inspect photovoltaic installations based on visual and thermal infrared analysis [6]. According to the obtained results, the inspection was done successfully. The obtained errors being at centimetric and millimetric level.

Drones will optimize surveillance costs and save time, increase performance and quality of monitored pipelines. Such services can be assured for system users [9]. The assessment of big cargo oil storage tanks can also be done faster by using drones than with manual methods [10]. Another advantage is given by the fact that the personnel is also put less at risk [11].

## 3 INSPECTION STRUCTURE

Pipelines are subject to corrosion and failure, as well as incidents and vandalism. Consequently, pipeline networks require periodic monitoring for maintenance, safety and security. According to this, damage assessment in transmission pipelines is an important factor and drones can help companies at the beginning of the occurrence of damage and allow experts to diagnose and plan maintenance. The use of high performance and resolution cameras supported by drones is a good method that evaluates pipes without any human contact and carries out rapid vision [3, 11-13]. The involvement of drones is useful for discovering potential failure in pipelines by sending a UAV into an area that would be dangerous for people. Inspectors can gather visual data about an asset condition via cognitive edge computing, artificial intelligence [14], automatic allocation for virtual machines [15, 16], and based on communication protocols between network nodes and address exchange [17].

The inspection basic process includes four main steps (Fig. 1): identifying pipeline path, identifying flight path, capturing the image, and processing (or post-processing).

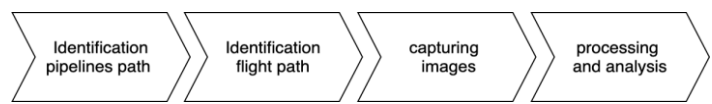

Figure 1. Inspection workflow steps

Each mission demands a plan to calculate all affecting factors, as well as a flight path. Different images are captured in various positions during the flight, then, the analysis process is done by the main software finding common features in images within the considered database. The accuracy is affected by two main factors, camera resolution and lens focal length. A camera with high resolution can show results in smaller pixel size, therefore it allows showing more smaller details.

## 4 RESULTS

The proposed system is analyzed according to a quantitative research method to trigger the opinion of the possible users and to determine the user requirements. 77 persons who work in the industry of oil and gas answer our survey which consists of 24 questions. The survey is distributed to persons who occupy an expert position in the oil or gas industry. According to the survey, 58.9% of the users believe that a drone inspection consists of a professional system that collects comprehensive visual data and identifies any potential obstacles and risks (Fig. 2). The other plausible drone inspections include detecting defects and performing land inspections.

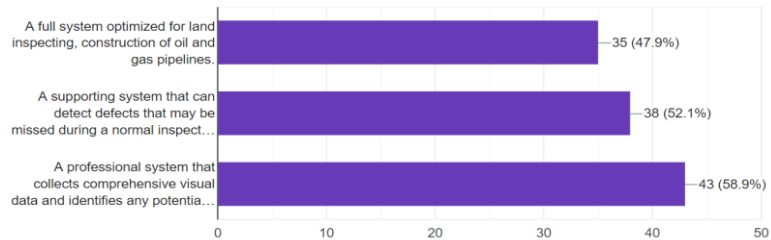

*Figure 2. Definition of drone inspection*

The users are confident that drones shape the future of inspections by flying over large areas and hard-to-reach properties (25.7%), provide technicians with more accurate visual data about pipe conditions (6.8%), improve maintenance work and increase productivity (4%) (Fig. 3). Out of all of them, 63.5% of users believe that all the previous mentioned characteristics can be achieved via drone inspection. Therefore, by providing visual inspection via the captured videos, the workers from the oil and gas industry will be helped to detect in a shorter amount of time the problems which exist.

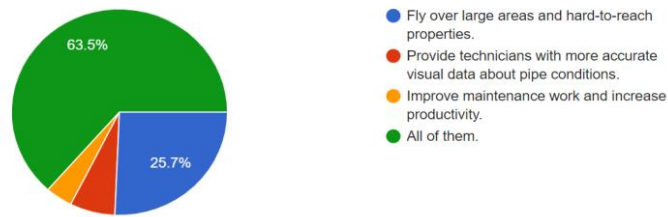

*Figure 3. Drone inspection types*

When asked about the improvement of safety (see Fig. 4), an inspection of toxic gases and chemicals occupies the first objective (75.7%), because it helps workers to bypass a series of health complications, such as cancer, decreased immunity, along with neurological and liver problems. Not only human health but also environment gets impacted due to oil and gas drilling, soil, air and water.

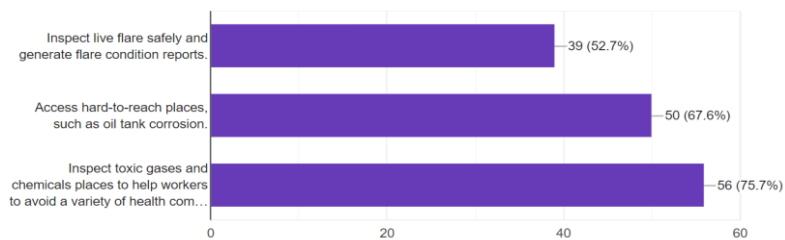

*Figure 4. UAV safety measures*

The potential of UAVs (Fig. 5) can be quickly exploited by providing erosion, ground displacement and vegetation monitoring activities (59.7%), monitoring and reporting pipe laying processes (54.2%), along with planning and preparing for trench digging and construction sites (48.6%). It is not very likely that sensors can be placed in the monitored site to detect leakage or to use the UAV to monitor oil field safety without exposure to the dangers of oil sites, namely pollution and radiation, including the security of dangerous sites.

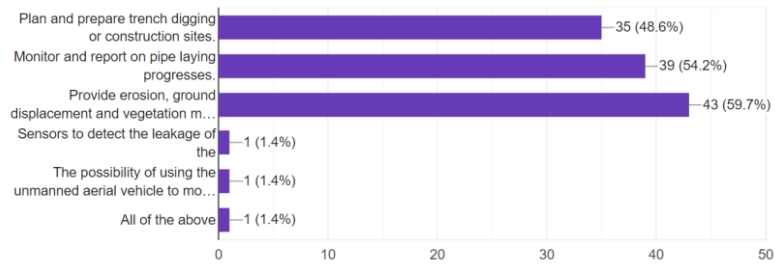

*Figure 5. UAV quick exploitation directions*

The future uses of drones in the oil and gas industry have also been analyzed, as in Fig. 6. 60.9% of the persons who answered the questionnaire believe that drones can provide pipeline moving operations, along with rig operations (53.6%). In this way, drones can transfer, handle and deliver materials (58%). In this way, critical situations can be solved in a shorter period of time and with fewer human resources, while improving their lives and being less exposed to risks. It was less likely believed that drones can provide surveillance to protect in case of crises, riots and wars (1.4%). 29.7% of persons are 100% sure that drones can overcome the challenges faced in the oil and gas industry.

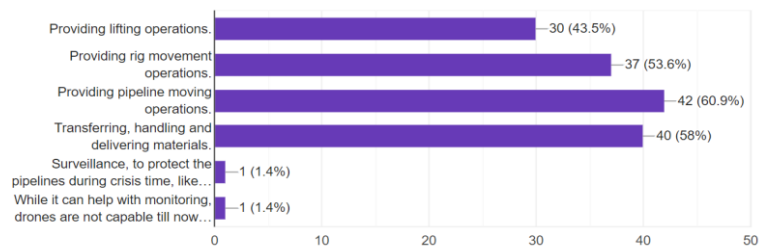

*Figure 6. Drone future uses*

The pipeline cases which drones can detect are upper ground pipelines (52.8%), followed by underground pipelines (6.9%), as in Fig. 7. It also included the case when leaks below ground can also be detected by drones (5.6%). 34.7% of the persons believe that all the previously mentioned cases can be detected via the use of drones.

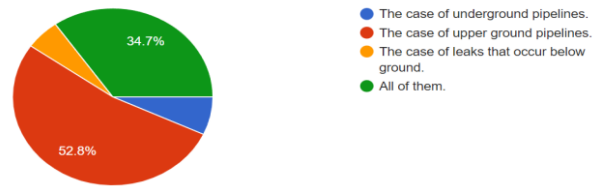

*Figure 7. Drone detection cases*

Drone-based inspections can include a combination of visual and thermography analysis. Based on this (see Fig. 8), users agree upon the fact that drones can determine a series of potentially hazardous problems before they become dangerous (29.2%), damage which can emerge as weakening of external insulation cladding (16.7%), along with piping thickness due to external corrosion (9.7%). 44.4% of the users agree that all the previously mentioned inspection types can be discovered in a short period of time by involving drones on a daily basis. The experts from the oil and gas industry assume that the higher altitude point to inspect assets safely and clearly by involving drones is 100 feet (38.4%), 200 feet (32.9%), 300 feet (19.1%) and 400 feet (9.6%), as in Fig. 9.

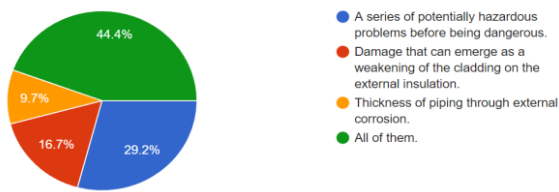

*Figure 8. Drone-based inspections*

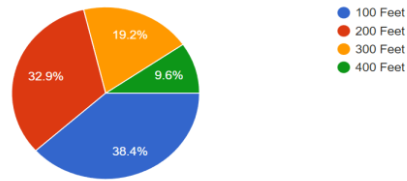

*Figure 9. High altitude drone inspection point*

The architecture of a drone-based system designed for the oil and gas industry involves several components, namely smartphone (47.2%), drone (61.1%), oil and gas pipelines (55.6%) and a system server (66.7%), as in Fig. 10.

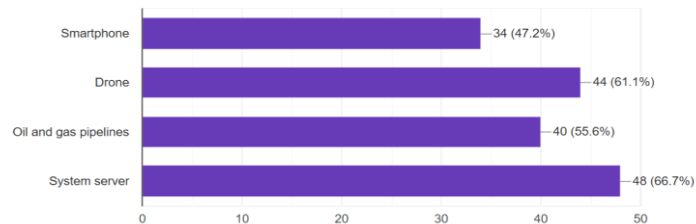

*Figure 10. Drone system architecture components*

Regarding the non-functional requirements of the drone-based system, the performances which should be achieved by using drones (see Fig. 11) are safety features for one drone to scan a territory, fly during the day, and avoid flying during cloudy weather (75.3%). Secondly, fast computations for the software application to determine critical situations are also needed (60.3%).

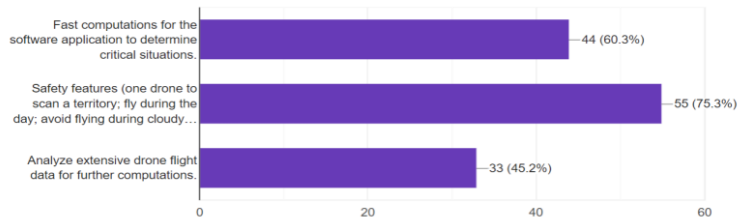

*Figure 11. Drone performance*

The availability rate which the drone system should offer is 99.999% according to 26.4% users, 90% based on 37.5%, 75% for 29.2% users and the rest desire a 50% rate (see Fig. 12).

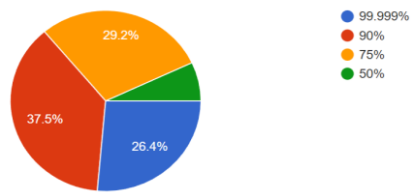

*Figure 12. Drone system availability rate*

The drone-based system should be accessible from the computer (79.2%), tablet (68.1%) and from mobile phones (62.5%), as illustrated in Fig. 13. The challenges that drones are facing are airspace concerns (78.1%), crashes (71.2%) and privacy (47.9%) (see Fig. 14).

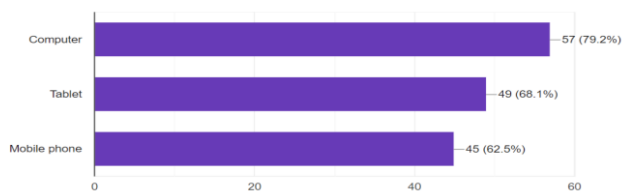

*Figure 13. Drone system accessibility*

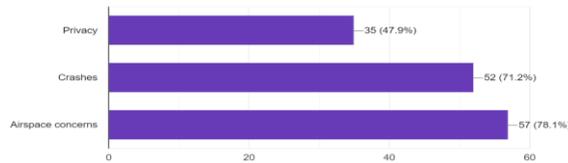

*Figure 14. Drone challenges*

The extensibility features which the system should offer are new functionalities that can be added inside the system (66.7%), response time which should take a short amount of time (61.1%), new user requests which can be incorporated (45.8%), along with managing scalability (38.9%) (see Fig. 15).

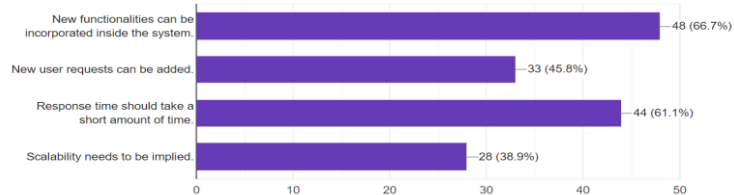

*Figure 15. Drone system extensibility features*

The security features (see Fig. 16) to be added to the system are secured authentication for users and administrators (72.6%), return home mode for drones to come back to the departure position if the battery is dying or the signal is jammed (65.8%). Other cases which need to be considered are drone firmware regular updates (49.3%), strong password usage (47.9%), virtual private network subscription (47.9%) and limitation for one device to connect to the system server (41.1%).

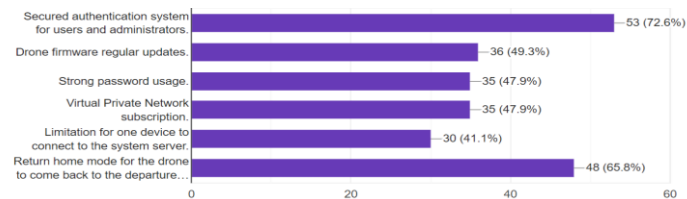

*Figure 16. Drone system security features*

The portability features which are needed for the system are for the application to run on different operating systems (71.4%), the software solution to be accessible via the Internet (61.9%) and for the application to run on different browsers (48.6%), as illustrated in Fig. 17. The transmission of data within the system is done such that drones inspect the oil and gas pipelines based on cameras and thermographic sensors (74.6%), drones transfer the captured data to the system server (67.6%) and data is stored at the server-side such that it can be analyzed via the same dedicated smartphone application (46.5%), as illustrated in Fig. 18.

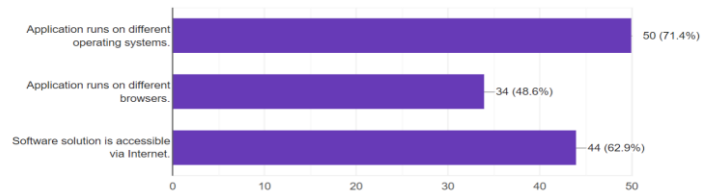

*Figure 17. Drone system portability features*

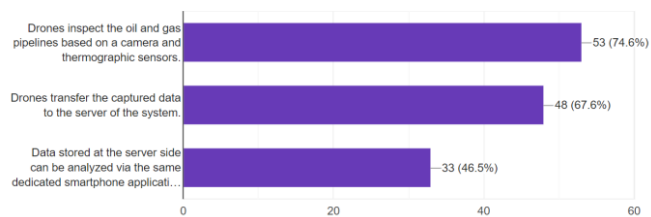

*Figure 18. Drone system data transmission*

In the case when the drone would lose control through the mission (see Fig. 19), the drone would automatically go back to the start point (40.3%), the drone would notify the server and send its coordinates (47.2%), and it would also notify the users of the area which they monitor (12.5%).

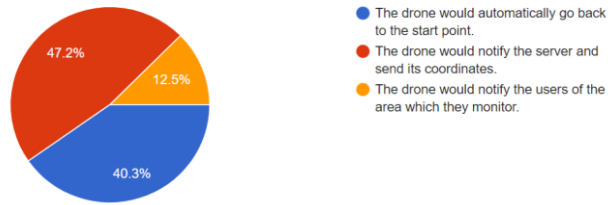

*Figure 19. Drone actions when control is lost*

## 5 CONCLUSIONS

The huge network of transmission pipelines creates a real challenge to oil and gas companies. The suggested inspection system, with the outline detection methods, permits the identification of large and small leaks for pipelines placed above or underground. The aerial monitoring system allows the procurement of high-resolution data at noticeably low costs and time.

The proposed solution can help workers from oil and gas universities, as well as it can avoid overburdening the considered pipeline system due to the dangerous outcomes related to pipe status. The solution can also be useful to professionals who work in the same domain. The current paper draws attention to the changes which may occur in a system where drones are used. The overall risks and costs can be diminished. Inspections based on drones can be extended to offer surveillance for wind turbines, roads, plants, tracks, industrial factories, power lines, health and safety. As future work, we would like to test in reality our described solution and to analyze the obtained results. The obtained error values can be diminished by enhancing our solution based on long-period tests. According to the online survey, the functional and non-functional requirements are outlined in the Table 2.

*Table 2. Functional and non-functional requirements*

| Functional Requirements (F) | Non-Functional Requirements (NF) |
|---|---|
| **F1.** The system is able to detect specific types of damages in transmission pipeline networks. | **NF1.** Performance |
| **F2.** The system uses a user interface to control the duties of drones. | **NF2.** Availability |
| **F3.** The onboard system is able to communicate with the ground station to receive or deliver updates. | **NF3.** Accessibility |
| **F4.** The system takes coordinates as input from the users with flight duration up to 3 hours. | **NF4.** Extensibility |
| **F5.** The system is able to get an altitude of a maximum of 400 feet. | **NF5.** Security |
| **F6.** The system has a professional camera to capture images and real-time video transmission with manual control during a mission. | **NF6.** Portability |

## ACKNOWLEDGEMENTS

The present article was financed by the University Politehnica of Bucharest, Romania, through the project "Inginer în Europa", in the online system, registered at ME under no. 140/GP/19.04.2021.